\documentclass{article}
\usepackage[nonatbib, final]{neurips_2024}
\usepackage[numbers]{natbib}
\usepackage[utf8]{inputenc} 
\usepackage[T1]{fontenc}    
\usepackage{hyperref}       
\usepackage{url}            
\usepackage{booktabs}       
\usepackage{amsfonts}       
\usepackage{nicefrac}       
\usepackage{microtype}      
\usepackage[dvipsnames,table,xcdraw]{xcolor}
\usepackage{amsmath}
\usepackage[pdftex]{graphicx}
\usepackage{multirow}
\usepackage{wrapfig}
\usepackage{caption}
\usepackage{enumitem}

\RequirePackage{algorithm}
\RequirePackage{algorithmic}

\newcommand{\wrt}{w.r.t. }
\newcommand{\nuq}{$\#$unique }
\newcommand{\ourmethodname}[1]{SED#1}

\newcommand{\uscore}{PDS}

\newcommand{\qualitativeimagesize}[1]{0.16#1}
\newcommand{\codeurl}[1]{\url{https://github.com/AlexanderRubinstein/diverse-universe-public}#1}

\title{Scalable Ensemble Diversification \\ for OOD Generalization and Detection}

\author{
  Alexander Rubinstein \\
  Tübingen AI Center, University of Tübingen\\
  \texttt{rubalex14@gmail.com} \\
  \And
  Luca Scimeca\phantom{aaaaa}\\
  Mila -- Quebec AI Institute\phantom{aaaaa}\\
  \texttt{luca.scimeca@mila.quebec}\phantom{aaaaa}\\
  \And
  \phantom{aaaaaaaa}Damien Teney \\
  \phantom{aaaaaaaa}Idiap Research Institute\\
  \phantom{aaaaaaaa}\texttt{damien.teney@idiap.ch} \\
  \And
  Seong Joon Oh \\
  Tübingen AI Center, University of Tübingen \\
  \texttt{coallaoh@gmail.com}
}

\begin{document}

\maketitle

\begin{abstract}
Training a diverse ensemble of models has several practical applications such as providing candidates for model selection with better out-of-distribution (OOD) generalization, and enabling the detection of OOD samples via Bayesian principles.
An existing approach to diverse ensemble training encourages the models to disagree on provided OOD samples.

However, the approach is computationally expensive and it requires well-separated ID and OOD examples, such that it has only been demonstrated in small-scale settings.

\textbf{Method.}
This work presents a method for Scalable Ensemble Diversification (\ourmethodname)
applicable to large-scale settings (e.g.\ ImageNet)
that does not require OOD samples.
Instead, \ourmethodname{} identifies hard training samples on the fly
and encourages the ensemble members to disagree on these.
To improve scaling, we show how to avoid the expensive computations in existing methods of exhaustive pairwise disagreements across models.

\textbf{Results.}
We evaluate the benefits of diversification with experiments on ImageNet.
First, for OOD generalization, we observe large benefits from the diversification in multiple settings including output-space (classical) ensembles and weight-space ensembles (model soups).
Second, for OOD detection, we turn the diversity of ensemble hypotheses into a novel uncertainty score estimator that surpasses a large number of OOD detection baselines.
\footnote{Code available at \codeurl{}.}
\end{abstract}

\section{Introduction}
Training an ensemble of diverse models is useful in multiple applications. Diverse ensembles are used to enhance out-of-distribution (OOD) generalization, where strong spurious features learned from the in-domain (ID) training data hinder generalization~\citep{DivDis, A2D, Damien1, Damien2}. By learning multiple hypotheses, the ensemble is given a chance to learn more predictive features that may otherwise be overshadowed by prominent non-robust and spurious features~\citep{chen2024project, yashima2022feature}. In Bayesian machine learning, diversification of the posterior samples has been studied as a means to improve the precision and efficiency of sample uncertainty estimates~\citep{d2021repulsive, MultiSwag2021}.

\begin{wrapfigure}{r}{.6\linewidth}
    \centering
    \includegraphics[width=1\linewidth]{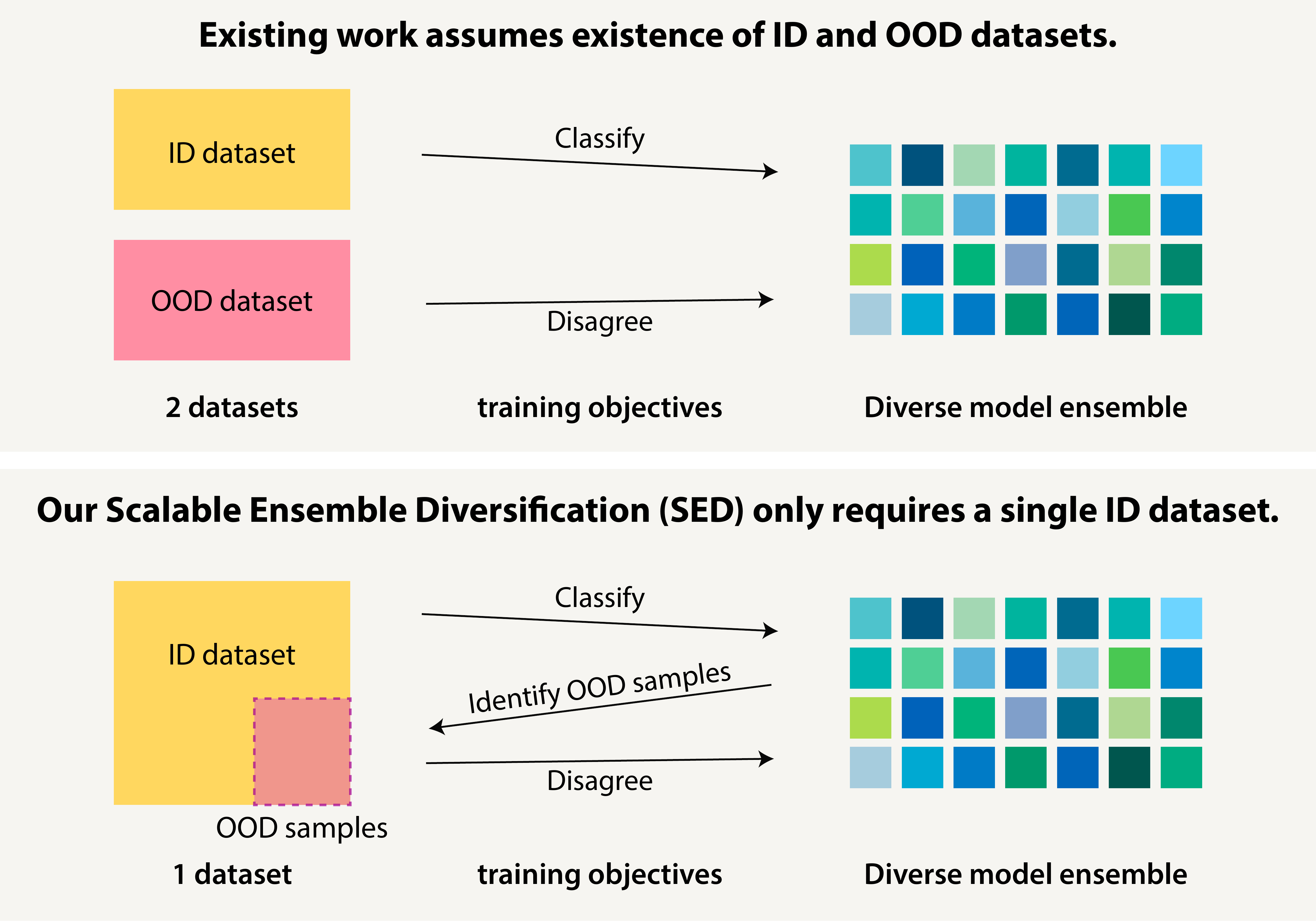}
    \caption{\small
    \textbf{Existing diversification methods (top)} require distinct (unlabeled) OOD examples on which the models are encouraged to disagree. Our \textbf{Scalable Ensemble Diversification (SED, bottom)} instead encourages the models to diverge on hard examples identified within a single standard training set.}
    \label{fig:teaser}
\end{wrapfigure}

A common strategy to train a diverse ensemble is to introduce a diversification objective while training the models in the ensemble in parallel~\citep{d2021repulsive, DivDis, A2D, ross2020ensembles, scimeca2023shortcut}. The main loss (e.g.\ cross-entropy for classification)
encourages the models to fit the labeled training,
while the diversification loss encourages the models to disagree with one another on unlabelled OOD samples~\citep{DivDis, A2D} (Figure~\ref{fig:teaser}).
The models are thus driven to discover different hypotheses that all explain the in-domain (ID)  data but behave different out of distribution.

The above approaches to diversification rely on the availability of two distinct sets of data:
labeled in-domain (ID) examples for the main training objective
and unlabeled OOD examples for diversification.

The existing methods are moreover computationally expensive, and have thus only been tested on small-scale artificial settings where the data can be clearly delineated into ID and OOD sets~\citep{DivDis, A2D}.
Some attempts were made to generate OOD data for diversification synthetically~\citep{scimeca2023shortcut}.
It is however still unclear how to apply these methods to realistic large-scale applications (e.g.\ ImageNet scale) where distinct OOD samples are not readily available.

\setlength{\intextsep}{0.5em}
\setlength{\columnsep}{0.7em}

This paper presents a method for Scalable Ensemble Diversification (\ourmethodname, Figure \ref{fig:teaser}) that addresses the limitations of existing approaches.
Starting with the ``\textit{Agree to Disagree}'' (A2D) method~\citep{A2D}, we introduce three technical innovations. (1)~Our method dynamically identifies hard samples from the training data on which the models are encouraged to disagree. (2)~At each iteration, the diversification objective is applied only on a random pair of models, alleviating the computational cost of the exhaustive pairing from prior work. (3)~The diversification objective is applied to deep networks by only affecting a small subset of layers towards the end of the network, further reducing the computational costs.
Altogether, these innovations enable scaling up to realistic applications that were so far out of reach for this family of methods.

Our experiments evaluate \ourmethodname{} by training a diverse ensemble on ImageNet.
We examine the benefits of the diversification for OOD generalization and OOD detection.
For OOD generalization, we showcase the usage of \ourmethodname-diversified ensemble in three variants: (a)~a classical ensemble (average of prediction probabilities)~\citep{Lakshminarayanan2017}, (b)~a model soup (average of model weights)~\citep{soup22}, and (c)~an oracle selection of the best individual model within the ensemble for each OOD test set~\citep{DivDis, Damien1}. In all three cases, \ourmethodname{} achieves superior generalization on multiple OOD test sets (ImageNet-A/R/C). For OOD detection, we examine multiple ways to use the \ourmethodname-diversified ensemble: (a)~treating them as samples of the Bayesian posterior and (b)~using a novel OODness estimate of Predictive Diversity Score (PDS) that measures the diversity of predictions from an ensemble. We show that PDS provides a superior detection of OOD samples like ImageNet-C, OpenImages, and iNaturalist.

Our contributions are summarized as follows.
\begin{enumerate}[itemsep=3pt,topsep=-3pt,partopsep=0pt,parsep=0pt,leftmargin=0.5cm]
    \item A novel method for Scalable Ensemble Diversification (\ourmethodname) that enables scaling up a popular approach to ensemble diversification based on prediction disagreement.
    \item A novel Predictive Diversity Score (\uscore) that estimates sample-wise OODness  based on ensemble prediction diversity.
    \item An empirical demonstration of ensemble diversification at the ImageNet scale, with demonstrated benefits in OOD generalization and detection.
\end{enumerate}

\section{Diverse Ensembles through Prediction Disagreement}

\textbf{Setting.}
We denote our training data $\mathcal{D} := \{x_{n}, y_{n}\}_{n=1}^N$ and refer to it as the in-domain (ID) data.
Prior diversification methods based on ``prediction disagreement''~\cite{DivDis, A2D}
require a separate set of unlabeled out-of-distribution (OOD) examples
$\mathcal{D}^\text{ood}:= \{x_{n}^\text{ood}\}_{n=1}^{N^\text{ood}}$.
Our proposed method will show how to proceed without $\mathcal{D}^\text{ood}$.
We denote with  $f(\cdot, \theta)$ a neural network classifier of parameters $\theta$. Then $f\left(x; \theta\right)\in\mathbb{R}^C$ corresponds to the logits over $C$ classes for the input $x$, and $p(x):= \operatorname{Softmax}(f(x))\in[0,1]^C$ probabilities over the classes.
Our goal is to obtain an ensemble $\{f^1,\cdots,f^M\}$ of $M$ models.
Our experiments in Section~\ref{sec:experiments} will showcase multiple ways to exploit these models (output-space ensembles, weight-space ensembles, etc.).

\textbf{Diversification through Disagreement.}
We now review the existing approach named ``\textit{Agree to Disagree}'' (A2D)~\citep{A2D} that we will improve upon in Section~\ref{sec:sed}.
The method trains the set of models $\{f^m\}_{m=1}^{m=M}$ in parallel with a main standard objective
and a diversification regularizer. 
The main objective is typically the cross-entropy loss
over all $M$ ensemble members and $N$ training examples:

\begin{align}
    \mathcal{L}_\text{main}=\frac{1}{MN}\sum_n\sum_m-\log p_{y_n}^m(x_n;\theta).
\end{align}
This encourages each member of the ensemble to similarly fit the training data.
The A2D diversification objective encourages a pair of models
$(f^m,f^l)$
to disagree (i.e.\ make different predictions) on OOD samples from $\mathcal{D}^\text{ood}$:
\begin{align}
    \label{eq:a2d_reg}
    \operatorname{A2D}\big( p^m(x), \,p^l(x) \big)
    ~=~\, -\log\!\Big(\,
    p^m_{\hat{y}}(x) \!\cdot\! \big(1 - p^l_{\hat{y}}(x)\big)
    ~+~
    p^l_{\hat{y}}(x) \!\cdot\!
    \big(1 - p^m_{\hat{y}}(x)\big)
    \,\Big)
\end{align}
where $\hat{y}:= \arg\max_c p^m_c(x)$ is the class predicted by the model $p^m$
(the definition could just as well use the prediction from $p^l$, which would make no practical difference~\citep{A2D}).
Minimizing (\ref{eq:a2d_reg}) encourages $p^l$ to assign a lower likelihood to the class predicted by $p^m$

and vice versa.
This is applied to all pairs of models from the ensemble and all OOD examples from $\mathcal{D}^\text{ood}$:

\begin{align}
\label{eq:A2D-on-OOD}
    \mathcal{L}_\text{div}=\frac{1}{N^\text{ood} M(M-1)}
    \sum^{N^\text{ood}}_{n=1}
    \sum_{m=1}^M
    \sum_{l=1}^{(m-1)}
    \operatorname{A2D}\big(p^m(x_n^\text{ood}), \, p^l(x_n^\text{ood})\big).
\end{align}

\section{Proposed Method}
\label{sec:sed}

We now present the Scalable Ensemble Diversification (\ourmethodname) method.
It improves upon A2D~\cite{A2D} by eliminating the need for OOD data and improving its computational efficiency.
The two technical novelties are the dynamic selection of hard samples within the training data (\S\ref{sec:dynamic-selection}) and the stochastic application to random pairs of models during training (\S\ref{sec:stochastic-pair-selection}).

\subsection{Dynamic Selection of Hard Examples}
\label{sec:dynamic-selection}

With no OOD data, it is difficult to apply disagreement methods since all models are being trained to fit all available training examples, i.e. to \emph{agree}.
Yet in practice, such OOD that that clearly differs from the ID data may not be readily available. It is not even clear how to define and obtain such OOD data where 
the feature-label correlations clearly differ from e.g.\ ImageNet.

To address this challenge, we propose to replace the OOD ``disagreement data''
with a set of hard training examples identified dynamically during training.
The models are then encouraged to disagree on these examples.
The desiderata for these hard samples are twofold: (a)~we wish to discriminate samples where the ensemble members make mistakes and (b)~we only trust the ensemble prediction for the hard sample identification when the ensemble is sufficiently trained. 

We assign a sample-wise weight $\alpha_n$ to each training sample $(x_n,y_n)\in\mathcal{D}$:
\begin{align}
    \label{eq:per_sample_budget}
    \alpha_n:= \dfrac{\text{CE}(f^1,\cdots,f^M;x_n,y_n)}{\left(\frac{1}{|B|}\sum_{b\in B} \text{CE}(f^1,\cdots,f^M;x_{b},y_{b})\right)^{2}}
\end{align}
where $\text{CE}(f^1,\cdots,f^M;x_n,y_n):=\text{CE}(\frac{1}{M} \sum_m f^m(x_n), y_{n})$ is the loss on the logit-averaged prediction and $B$ is a mini-batch that contains the sample $(x_n,y_n)$. $\alpha_n$ is a weight proportional to the ensemble loss on the sample, which fulfills desideratum~(a) mentioned above.
The normalization then handles desideratum~(b). To see this, consider the batch-wise weight:
\begin{align}
    \alpha_{B}:= \frac{1}{|B|}\sum_{b\in B} \alpha_b = \frac{1}{\frac{1}{|B|}\sum_b \text{CE}(f^1,\cdots,f^M;x_b,y_b)}.
\end{align}
Now $\alpha_{B}$ is \textit{inversely proportional} to the average cross-entropy loss of the ensemble on the mini-batch $B$. Thus, the overall level of $\alpha_{n}$ for $n\in B$ is lower for earlier iterations of the ensemble training, where the predictions from the models are not trustworthy yet.

We now use the sample-wise weights $\alpha_n$ to define the SED training objective:
\begin{align}
\label{eq:sed}
\mathcal{L}_\text{SED} ~:=~ \mathcal{L}_\text{main} + \frac{\lambda}{N M(M-1)}\sum_n\sum_{m<l} \operatorname{stopgrad}(\alpha_n) \cdot \operatorname{A2D}\big(p^m(x_n), p^l(x_n)\big),
\end{align}
where $\lambda>0$ controls the strength of the diversification. The operator $\operatorname{stopgrad}(\cdot)$ outputs a copy of its argument that is treated as a constant during backpropagation.
Compared to Equation \ref{eq:A2D-on-OOD}, this formulation does not require OOD disagreement data.
Instead, all training examples are treated as potential hard samples to disagree on, and their difficulty is softly determined via $\alpha_n$. 

Most of our experiments use A2D as the diversity regularizer because it is considered state-of-the-art~\citep{benoit2024unraveling}, theoretically sound, and it performs well in our experiments. Alternative diversity regularizers could be used such as DivDis~\citep{DivDis} as demonstrated for comparison in Table~\ref{tab:ood_abl}.

\subsection{Tricks to Improve Scalability}
\label{sec:stochastic-pair-selection}

\begin{wrapfigure}{r}{0.3\textwidth}
    \centering
    \vspace{-1em}
    \includegraphics[width=1\linewidth]{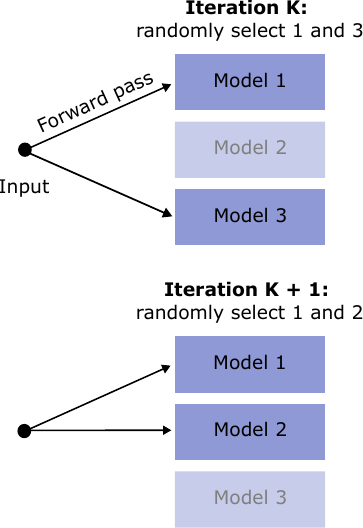}
    \vspace{-1em}
\end{wrapfigure}

Many diversification algorithms are based on exhaustive pairwise comparisons between all the models in the ensemble (see the second term of Equation \ref{eq:sed}). This scales quadratically with the size $M$ of the ensemble.

We propose to use a stochastic sum. For every mini-batch $B$, we use a random subset of models $\mathcal{I} \in \{1,\cdots,M\}$ on which to compute the diversification term in Equation \ref{eq:sed} (see figure on the right). In our experiments, we randomly sample one pair of models per batch ($\mathcal{I} = 2$). Interestingly, we noticed empirically that this stochastic sum sometimes induces diversity by itself (without a diversification term) and leads to better performance.

To further speed up the training, we consider updating only a subset of the layers of the model with the SED objective, keeping others frozen.
More specifically, each ensemble member in the experiments of Section~\ref{sec:experiments}
is based on a frozen Deit3b model~\citep{deit3_2022}
of which we diversify only the last two layers.

\subsection{Predictive Diversity Score (PDS) for OOD Detection}
\label{sec:epistemic_unc}

We now describe how to use diverse ensembles for OOD detection~\citep{helton2004exploration}.
This is based on evaluating the epistemic uncertainty, which is the consequence of the lack of training data in a given regions of the input space~\citep{mukhoti2023deep, hullermeier2021aleatoric}. In these OOD regions, the lack of supervision means that diverse models are likely to disagree in their predictions~\citep{malinin2019ensemble, DivDis, A2D}. 
We therefore propose to use the \emph{agreement rate} across models on given sample to estimate the epistemic uncertainty and its ``OODness''.

\textbf{BMA Baseline.}~~
Given an ensemble of models, a simple baseline for OOD detection is to compute the predictive uncertainty of the Bayesian Model Averaging (BMA) by treating the ensemble members as samples of the posterior $p(\theta | \mathcal{D})$~\citep{Lakshminarayanan2017, MultiSwag2021}:
\begin{align}
    \eta_\text{BMA}:=\max_c\frac{1}{M}\sum_m p^m_c(x).
\end{align}
While being a strong baseline~\citep{mukhoti2023deep} for OOD detection this notion of uncertainty does not directly exploit the potential diversity in individual models of the ensemble because it averages out the predictions along the model index $m$. In addition to that, mimicking the true distribution makes individual members have small values of $\max_c p^m_c(x)$ on training samples with high aleatoric uncertainty~\citep{hullermeier2021aleatoric}. This is why BMA is not a reliable indicator of epistemic uncertainty.

\textbf{Proposed Predictive Diversity Score (PDS).}~~
We propose a novel measure for epistemic uncertainty that directly measures the prediction diversity of the individual members. Concretely,
\begin{align}
    \eta_\text{PDS}:= \frac{1}{C}\sum_c \max_m p_{c}^{m}(x).
\end{align}
PDS is a continuous relaxation of the number of unique argmax predictions within an ensemble of models. To see this, consider the special case where $p^m\in\{0,1\}$ are one-hot vectors. Then, $\max_m p_{c}^{m}(x)$ is 1 if any of $m$ predicts $c$ and 0 otherwise. Thus, $\sum_c \max_m p_{c}^{m}(x)$ computes the number of classes predicted by at least one ensemble member.

\section{Experiments}
\label{sec:experiments}

We present experiments that first evaluate the intrinsic 
diversification from SED (\S\ref{sec:diversification}) then evaluate several use cases of diverse ensembles for
OOD generalization (\S\ref{sec:ood-generalization})
and OOD detection (\S\ref{sec:ood-detection}).

\subsection{Experimental Setup}
\label{sec:setup}

\textbf{Implementation.} For both tasks, we train an ensemble of models with SED based on A2D~\citep{A2D} using the AdamW optimizer~\citep{AdamW2019}, a batch size varies from $16$ to $256$, learning rate from $10^{-4}$ to $10^{-3}$, weight decay is fixed to $0.01$, and number of epochs to $10$. The diversity weight $\lambda$ varies from $0$ to $5$ and the stochastic pairing is done for $|\mathcal{I}|=2$ models for each mini-batch (see Table~\ref{tab:stoch_sum_abl} for an evaluation of other values). All experiments use models based on the Deit3b architecture~\citep{deit3_2022} pretrained on ImageNet21k~\citep{IN21k2009}. As suggested in \S\ref{sec:stochastic-pair-selection} we train  only the last $2$ layers. As in-domain (ID) data we use the training split of ImageNet ( $|\mathcal{D}|=1,281,167$). All experiments were run on RTX2080Ti GPUs with 12GB vRAM and 40GB RAM. Each experiment took between 2 to 12 hours.

\textbf{Baselines.} As a simple ensemble we use a variant of \textit{deep ensembles}~\citep{Lakshminarayanan2017}, which uses models trained independently with different random seeds.

To match the resource usage of our SED, we also train only the last 2 layers of the models (i.e.\ they are ``shallow ensembles'').

We also consider simple ensembles of models with diverse hyperparameters~\citep{wenzel2020hyperparameter}.
We reimplemented A2D~\citep{A2D} and DivDis~\citep{DivDis}, with which we use unlabeled samples from ImageNet-R as disagreement data (the choice of dataset used for disagreement has little influence on the results, as seen in Table~\ref{tab:ood_abl}). For A2D, we use a frozen feature extractor and parallel training, i.e.\ all models are trained simultaneously rather than sequentially.

\textbf{Evaluation of OOD generalization.}
We evaluate the classification accuracy of the ensembles trained on ImageNet with the (ID) validation split of ImageNet (IN-Val, 50,000 samples) and multiple OOD datasets: ImageNet-A (\textit{IN-A}~\citep{IN-A_IN-O_2021}, 7.5k images \& 200 classes), ImageNet-R (\textit{IN-R}~\citep{IN-R_2021}, 30k images, 200 classes), ImageNet-C (\textit{IN-C-i} or just \textit{C-i} for corruption strength $i$~\citep{IN-C_2019}, 50k images, 1k classes). OpenImages-O (\textit{OI}~\citep{wang2022vim}, 17k images, unlabeled), and iNaturalist (\textit{iNat}~\citep{huang2021mos}, 10k images, unlabeled).

\textbf{Evaluation of OOD detection.}
The task is to differentiate samples from the above OOD datasets against those from the ImageNet validation data (considered as ID). The evaluation includes both ``semantic'' and ``covariate'' types of shifts~\citep{zhang2023openood, IN-C_2019, IN-R_2021, recht2019imagenet, yang2024imagenetood}. Openimages-O and iNaturalist represent semantic shifts because their label sets are disjoint from ImageNet's. And ImageNet-C represents a covariate shift because its label set is the same as ImageNet's but the style of images differs. We measure the OOD detection performance with the area under the ROC curve, following~\citep{hendrycks2017a}.

\subsection{Diversification}
\label{sec:diversification}

We start with the question of whether SED truly diversifies the ensemble.
To measure the diversity of the ensemble, we compute the number of unique predictions for each sample for the committee of models ($\#$unique).

\begin{table}
\vspace{-.7em}
  \centering
  \setlength{\tabcolsep}{.5em}
  \small
\begin{tabular}{l|cc|cccc}
 \toprule
                   &   IN-Val& IN-Val&IN-C-1&IN-C-5& \multicolumn{1}{l}{iNaturalist} & \multicolumn{1}{l}{OpenImages} \\
 \midrule
Detector type& Covariate & Semantic & Covariate& Covariate& Semantic &Semantic \\
\midrule
Deep ensemble &  1.05 (1.10)& 1.05 (1.10)&1.09 (1.11)& 1.19 (1.11)& 1.31 (1.10)& 1.23 (1.10)\\
+Diverse hyperparams &  1.04 (1.17)& 1.04 (1.17)&1.11 (1.23)& 1.32 (1.37)& 1.48 (1.41)& 1.33 (1.38)\\
\midrule
A2D &  1.11 (1.15)& 1.11 (1.15)&1.04 (1.04)& 1.15 (1.10)& 1.19 (1.15)& 1.91 (1.49)\\
\midrule
\ourmethodname-A2D &  \textbf{5.00 (3.98)}& \textbf{1.36 (1.54)}&\textbf{5.00 (4.16)}& \textbf{5.00 (4.46)}& \textbf{4.68 (4.06)}& \textbf{4.11 (3.53)}\\
\bottomrule
 \end{tabular}

 \vspace{1em}
  \caption{\small\textbf{Diversity measure for ensembles.} We report the \nuq and PDS (in parentheses) on OOD datasets and IN-Val dataset (See \S\ref{sec:setup} for the datasets). The ensemble size $M$ is 5 for all methods; it is the max possible \nuq value.
  \textbf{Covariate/semantic shift detectors:} for \textit{Covariate} detector type we provide \nuq (PDS) of an ensemble with the best OOD detection performance on IN-C-1 and IN-C-5 while for \textit{Semantic} detector type we provide \nuq (PDS) of an ensemble with the best OOD detection performance on iNaturalist and OpenImages}
  \label{tab:diversity}
  \vspace{-2em}
\end{table}

Table~\ref{tab:diversity} shows the $\#$unique and PDS values for the IN-Val as well as multiple OOD datasets. We observe that the deep ensemble baseline does not increase the diversity dramatically (e.g.\ 1.09 for IN-C-1) beyond no-diversity values (1.0). Diversification tricks like hyperparameter diversification (1.11 for IN-C-1) or A2D (1.04 for IN-C-1) and DivDis (1.04 for IN-C-1) only marginally change the prediction diversity. On the other hand, our SED increases the prediction diversity across the board (e.g.\ 5.00 for IN-C-1). It is important to note that for covariate shift detector, an ensemble with the best OOD detection performance on covariate shift datasets (IN-C-1 and IN-C-5), \nuq is high for IN-Val dataset as well (5.00 in IN-Val Cov column). Nevertheless, it has the best performance in OOD detection for covariate shifts (IN-Val vs IN-C-1/IN-C-5) when using PDS as uncertainty score because its value is still lower for IN-Val than for OOD (3.98 vs 4.16/4.46 for IN-Val vs IN-C-1/IN-C-5).

\begin{figure}
  \centering
  \small
  \setlength{\tabcolsep}{.5em}
  \renewcommand{\arraystretch}{.7}
\begin{tabular}{llllll}
\toprule&\includegraphics[width=\qualitativeimagesize\linewidth]{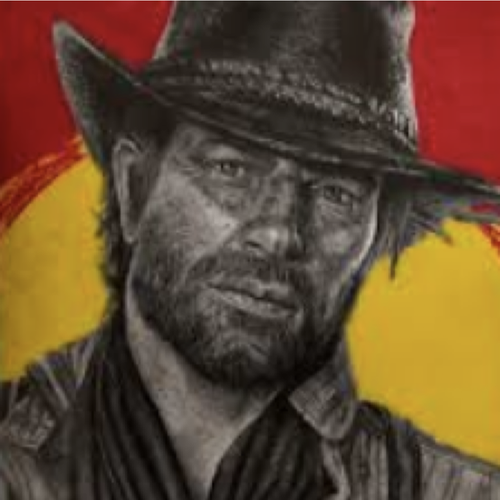}& \includegraphics[width=\qualitativeimagesize\linewidth]{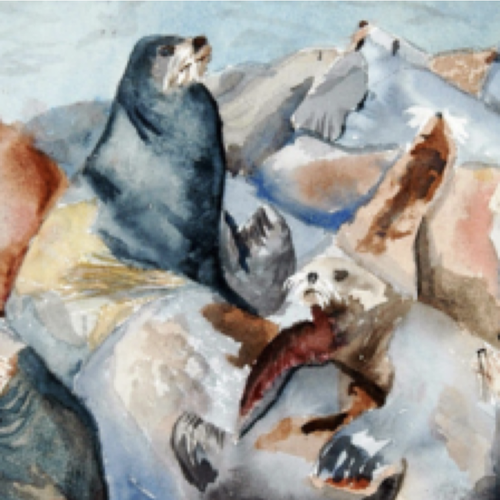}& \includegraphics[width=\qualitativeimagesize\linewidth]{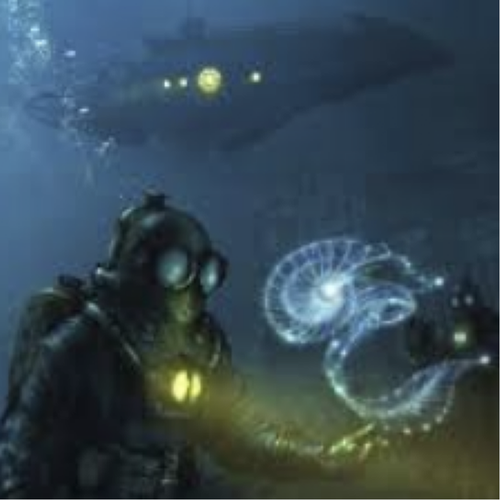}& \includegraphics[width=\qualitativeimagesize\linewidth]{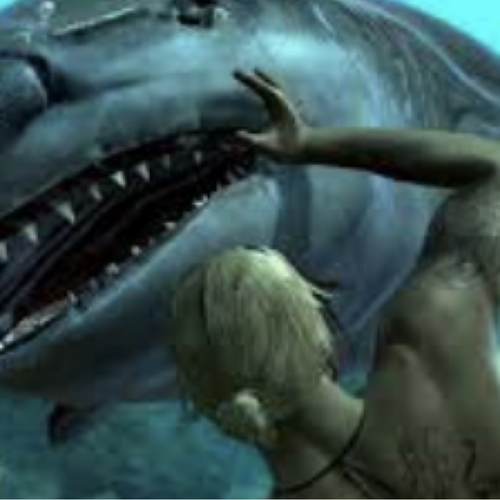}& \includegraphics[width=\qualitativeimagesize\linewidth]{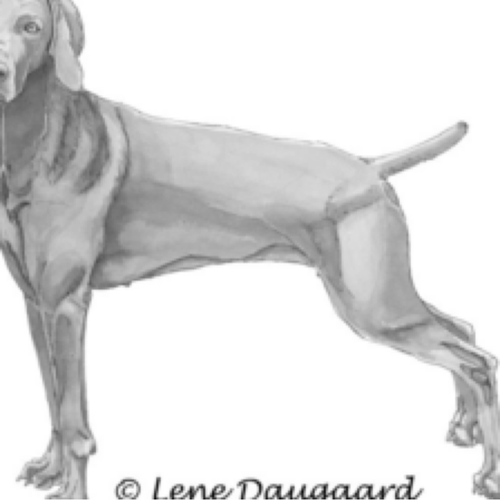}\\
                   GT&Cowboy hat&Sea lion&Scuba diver&Great shark& \multicolumn{1}{l}{Weimaraner}  \\

 \midrule
 \ourmethodname{}&{\textbf{Cowboy hat}}&{\textbf{Sea lion}}&{\textbf{Scuba diver}}&{\textbf{Great shark}}& {\textbf{Weimaraner}}\\
 & Comic book& Otter& Jellyfish& Killer whale&Vizsla\\
 \midrule
 \uscore&0.300& 0.300& 0.294& 0.292& 0.292\\
 \midrule
  &\includegraphics[width=\qualitativeimagesize\linewidth]{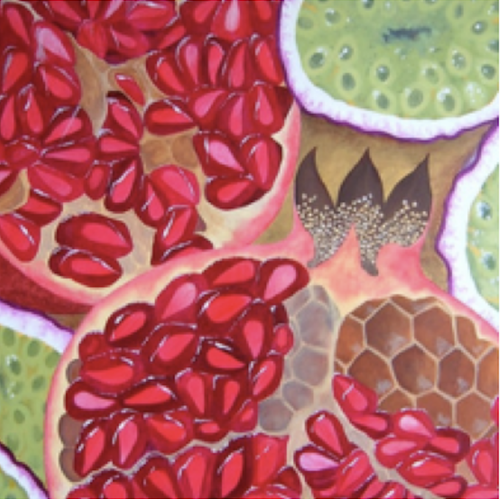}& \includegraphics[width=\qualitativeimagesize\linewidth]{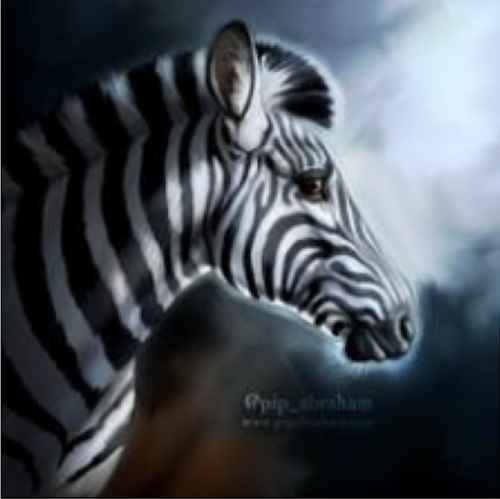}& \includegraphics[width=\qualitativeimagesize\linewidth]{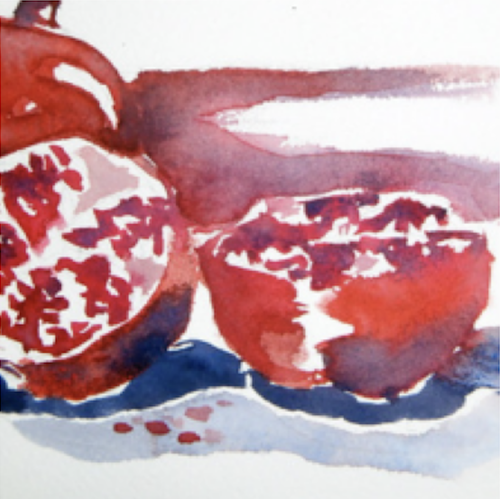}& \includegraphics[width=\qualitativeimagesize\linewidth]{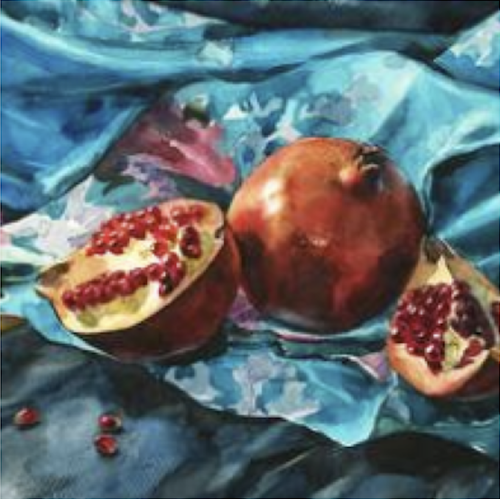}& \includegraphics[width=\qualitativeimagesize\linewidth]{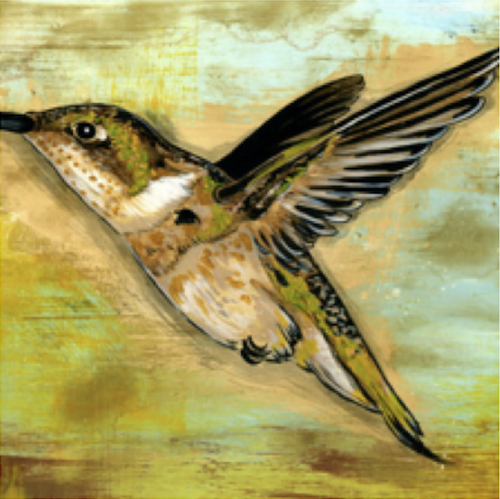}\\
 GT& Pomegranate& Zebra& Pomegranate& Pomegranate&Hummingbird\\
 \midrule
 \ourmethodname{}& {\textbf{Pomegranate}}& {\textbf{Zebra}}& {\textbf{Pomegranate}}& {\textbf{Pomegranate}}&{\textbf{Hummingbird}}\\
 \midrule
 \uscore& 0.216& 0.216& 0.216& 0.216&0.216\\
 \bottomrule
\end{tabular}
  \caption{\small\textbf{ImageNet-R examples leading to the greatest and least disagreement}.
  We show the 5 most divergent and 5 least divergent samples according to the SED ensemble. 
  We measure prediction diversity with the Prediction Diversity Score (PDS) in \S\ref{sec:epistemic_unc}. GT refers to the ground truth category. Ensemble predictions are shown in bold; in cases where ensemble members predict classes different from the ensemble prediction we provide them on the next line with standard font.}
  \label{fig:qual_div}
  \vspace{-1em}
\end{figure}

Qualitative results on ImageNet-R further verify the ability of SED to diversify the ensemble (Figure~\ref{fig:qual_div}). As a measure for diversity, we use the Predictive Diversity Score (PDS) in \S\ref{sec:epistemic_unc}. We observe that the samples inducing the highest diversity (high PDS scores) are indeed ambiguous: for the first image, where the ``cowboy hat'' is the ground truth category, we observe that ``comic book'' is also a valid label for the image style. On the other hand, samples with low PDS exhibit clearer image-to-category relationship.

\subsection{OOD Generalization}
\label{sec:ood-generalization}

We examine the first application of diverse ensembles: OOD generalization. We hypothesize that the superior diversification ability verified in \S\ref{sec:diversification} leads to greater OOD generalization due to the consideration of more robust hypotheses that do not rely on obvious spurious correlations. 

\textbf{Ensemble aggregation for OOD generalization.} 
As a means to exploit such robust hypotheses, we consider 3 aggregation strategies. (1) \textit{Oracle selection}: the best-performing individual model is chosen from an ensemble~\citep{A2D, Damien1}. The final prediction is given by $f(x;\theta^{m^\star})$ where $m^\star:=\arg\max _m\operatorname{Acc}(f^m, \mathcal{D}^\text{ood})$.
(2) \textit{Prediction ensemble} is a vanilla prediction ensemble where the logit values are averaged: $\frac{1}{M} \sum_m f^m(x)$~\citep{soup22}.
(3) \textit{Uniform soup}~\citep{soup22} averages the weights themselves. The final prediction is given by $f(x; \frac{1}{M} \sum_m \theta^m)$.

\textbf{SED improves OOD generalization for ensembles.}
We show the OOD generalization performance of ensembles in Table \ref{tab:ood-generalization}, for the three ensemble prediction aggregation strategies described above. We observe that our SED framework (SED-A2D) results in superior OOD generalization performance for the prediction ensemble and uniform soup while being on par with best baselines for the oracle selection. SED-A2D is particularly strong in the prediction ensemble (e.g.\ 48.7\% for $M=5$ and 53.8\% for $M=50$ on ImageNet-R) and uniform soup (e.g.\ 46.1\% for $M=5$ and 46.5\% for $M=50$ on ImageNet-R). We contend that the increased ensemble diversity contributes to the improvements in OOD generalization. We also remark that the SED framework (SED-A2D) envelops the performance of A2D in this ImageNet-scale experiment. Together with the superiority of computational efficiency (as discussed at the end of \S~\ref{sec:ood-detection}) of SED-A2D over the A2D, this demonstrates that SED fulfills its purpose of scaling up ensemble diversification methods like A2D.

\textbf{Deep ensembles are a strong baseline.}
We also note that deep ensemble, particularly with diverse hyperparameters, provides a strong baseline, outperforming dedicated diversification methodologies under the oracle selection strategy when $M=5$. It also provides a good balance between ID (ImageNet validation split) and OOD generalization. 

\begin{table}
  \setlength{\tabcolsep}{.25em}
  \centering
  \small
     \begin{tabular}{l|c|ccccc|ccccc|ccccc}
    \toprule
    \multicolumn{2}{c}{} & \multicolumn{5}{c}{Oracle selection} & \multicolumn{5}{c}{Prediction ensemble} & \multicolumn{5}{c}{Uniform soup} \\
    \midrule
     Method &   $M$ & Val & IN-A & IN-R & C-1 & C-5 & Val & IN-A & IN-R & C-1 & C-5 & Val & IN-A & IN-R & C-1 & C-5 \\
     \midrule
 Single model &  1 & 85.4 & 37.9 & 44.7 & 75.6 & 38.5 & 85.4 & 37.9 & 44.7 & 75.6 & 38.5 & 85.4 & 37.9 & 44.7 & 75.6 & 38.5 \\
    \midrule
    Deep ensemble & 5 & \textbf{85.4} & 37.9 & 44.9 & 75.7 & 38.6 & \textbf{85.4} & 39.9 & 46.3 & 75.7 & 38.6 & \textbf{85.3} & 36.7 & 44.6 & 75.5 & 38.3 \\
    +Diverse HPs & 5 & \textbf{85.4} & \textbf{38.5} & \textbf{45.4} & \textbf{77.4} & \textbf{40.7} & \textbf{85.4} & 39.9 & 46.5 & 76.0 & 39.0 & \textbf{85.3} & 35.3 & 44.1 & 75.9 & 38.7 \\
     DivDis & 5 & 85.2 & 35.8 & 40.8 & 77.2 & 40.2 & 85.1 & 36.3 & 41.8 & 77.2 & 40.2 & 84.8 & \textbf{40.7} & 42.5 & 76.2 & 38.9 \\
     A2D & 5 & 85.2 & 36.6 & 44.3 & 77.3 & 40.4 & 85.1 & 37.8 & 45.2 & 77.2 & 40.3 & 84.5 & 39.3 & 45.1 & 75.5 & 39.1 \\
     \ourmethodname-A2D & 5 & 85.1 & 38.3 & 45.3 & 77.2 & 40.4 & 85.3 & \textbf{43.0}& \textbf{48.7}& \textbf{77.3} & \textbf{40.7}& \textbf{85.3} & 40.3 & \textbf{46.1} & \textbf{77.3} & \textbf{40.6} \\
     \midrule
 Deep ensemble & 50 & \textbf{85.5} & 38.1 & 45.2 & 75.7 & 38.6 & \textbf{85.5} & 38.8 & 45.8 & 75.6 & 38.5 & \textbf{85.4} & 37.5 & 45.0 & 75.5 & 38.4 \\
 +Diverse HPs & 50 & 85.5 & 38.5 & 45.6 & \textbf{77.5} & \textbf{40.8} & 85.5 & 42.5 & 48.5 & \textbf{76.0} & 39.0 & \textbf{85.4} & 36.4 & 44.8 & \textbf{75.9} & 38.8 \\
 \ourmethodname-A2D & 50 & 82.6& \textbf{39.0}& \textbf{45.8}& 74.4& 38.3& 83.6& \textbf{50.6}& \textbf{53.8}& 75.8& \textbf{39.3}& 83.5& \textbf{39.2}& \textbf{46.5}& 75.8& \textbf{39.3}\\
 \bottomrule
 \end{tabular}
    \vspace{.5em}
\caption{\small\textbf{OOD generalization of ensembles.} Models are trained on the ImageNet training split. $M$ is the ensemble size. For DivDis and A2D, we use the ImageNet-R as the OOD datasets where the respective diversification objectives are applied.}
\label{tab:ood-generalization}
\end{table}

\subsection{OOD Detection}
\label{sec:ood-detection}

\begin{table}
  \centering
  \small
  \setlength{\tabcolsep}{.7em}
  \begin{tabular}{l|c|cc|cccc}
     \toprule
     & $\eta$ &   IN-Val&IN-Val &IN-C-1 & IN-C-5 & iNaturalist & OpenImages \\
     \midrule
    Detector type& & Covariate & Semantic& Covariate& Covariate& Semantic&Semantic\\
    \midrule
    Single model & BMA &   85.4&85.4&0.615& 0.833& 0.958& 0.909\\
    \midrule
    Deep Ensemble & BMA &   85.5&85.5&0.619 & 0.835 & 0.958 & 0.911\\
    +Diverse HPs & BMA &   85.5&85.5&\textbf{0.642} & \textbf{0.861} & \textbf{0.969} & \textbf{0.923}\\
    DivDis & BMA &   85.2&85.2&0.598 & 0.843 & 0.966 & 0.922\\
    A2D & BMA &   84.7&85.2&0.594 & 0.835 & 0.966 & 0.916\\
    \ourmethodname{}-A2D & BMA &   85.1&77.5&0.641 & 0.845 & 0.960 & 0.915\\
    \midrule
    Deep Ensemble & PDS &   85.5&85.5&0.565 & 0.625 & 0.592 & 0.589\\
    +Diverse HPs & PDS &   85.5&85.5&0.643 & 0.849 & 0.926 & 0.889\\
    DivDis & PDS &   85.2&85.2&0.600 & 0.851 & 0.969 & 0.939\\
    A2D & PDS &   85.2&85.2&0.599 & 0.850 & 0.971 & 0.939\\
    \ourmethodname{}-A2D & PDS &   1.0&82.9&\textbf{0.681}& \textbf{0.894}& \textbf{0.977} & 
    \textbf{0.941}\\
    \bottomrule
\end{tabular}
\vspace{1em}
  \caption{\small\textbf{OOD detection via ensembles.} For each OOD dataset (IN-C-1, IN-C-5, iNaturalist, and OpenImages), the ensembles are tasked to detect the respective OOD samples among IN-Val samples (ImageNet validation split). We show the AUROC scores for the OOD detection task. To visualize the tradeoff between the classification and OOD detection tasks we also show the accuracy of the corresponding models on IN-Val samples. Covariate and Semantic detector types are the same as in Table~\ref{tab:diversity}. Ensemble size is fixed at $M=5$. $\eta$ refers to the epistemic uncertainty computation framework discussed in \S\ref{sec:epistemic_unc}.}
  \label{tab:ood_det}
  \vspace{-1em}
\end{table}

We study the impact of ensemble diversification on OOD detection capabilities of an ensemble. Once an ensemble is trained, we compute the epistemic uncertainty, or likelihood of the sample being OOD, following two schemes, $\eta_\text{BMA}$ and $\eta_\text{PDS}$ introduced in \S\ref{sec:epistemic_unc}.

\textbf{SED and PDS together lead to superior OOD detection performance.} 
We show the OOD detection results in Table~\ref{tab:ood_det}. We chose BMA because it is considered a standard baseline~\citep{mukhoti2023deep} for uncertainty quantification, comparison to other OOD detection methods can be seen in Table~\ref{tab:ood_scores}. For the BMA scores, deep ensemble remains a strong baseline. In particular, when the hyperparameters are varied (``+Diverse HPs''), the detection AUROC reaches the maximal performance among the ensembles using the BMA scores. The quality of PDS is more sensitive to the ensemble diversity, as seen in the jump from the deep ensemble (e.g.\ 0.589 for OpenImages) to the diverse-HP variant (0.889). However, when the ensemble is sufficiently diverse, such as when trained with SED-A2D, the PDS leads to high-quality OODness scores. SED-A2D with PDS achieves the best AUROC across the board, including the BMA variants. However, superior OOD detection performance comes at the cost of drop in classification to 82.9 for semantic shift detector and 1.0 for covariate shift detector (see Table~\ref{tab:diversity} for detector types).

\begin{figure}[!ht]
    \small
    \centering
    \setlength{\tabcolsep}{0em}
    \begin{tabular}{cc}
        \includegraphics[width=0.499\linewidth]{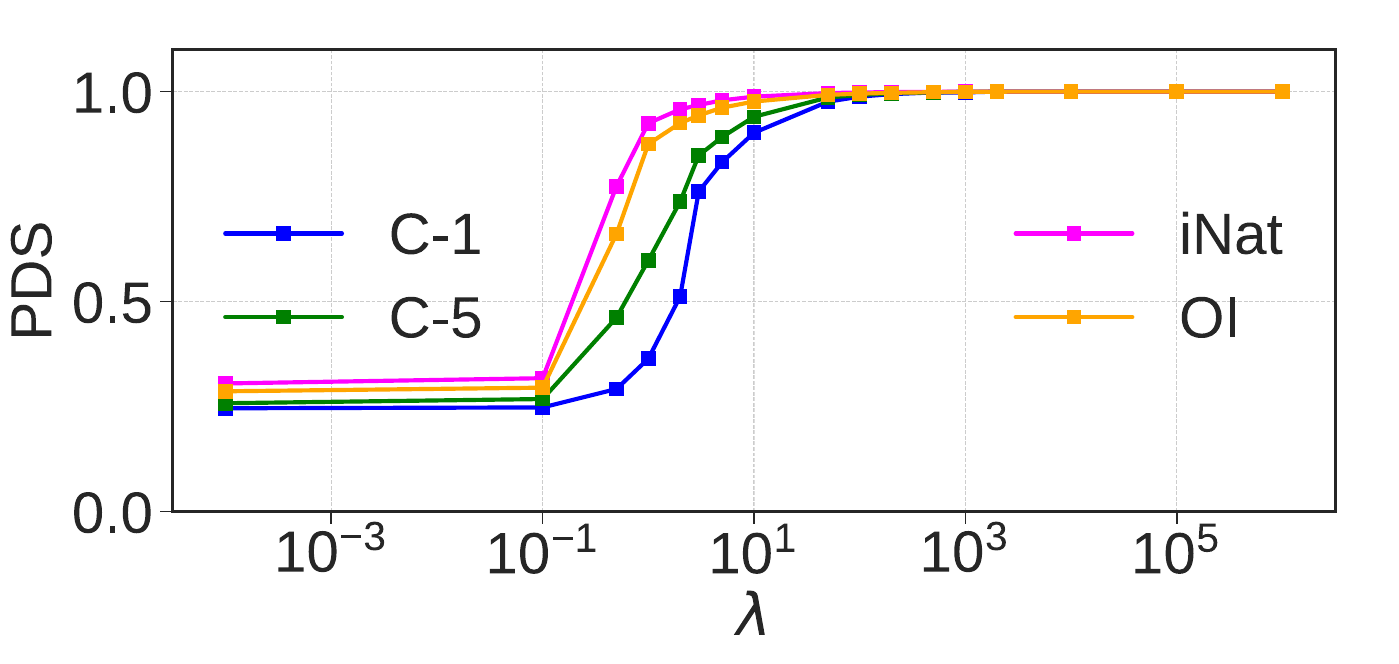}
        &
        \includegraphics[width=0.499\linewidth]{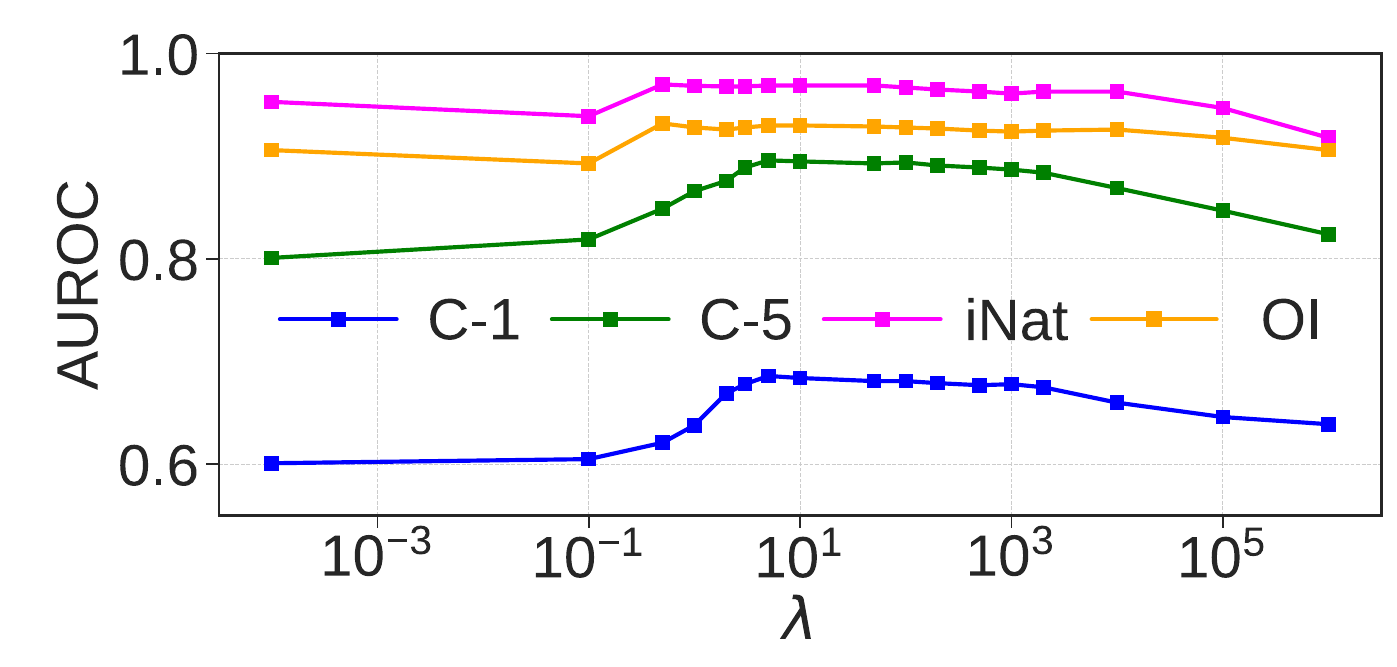}
    \end{tabular}
    \vspace{-1em}
    \caption{\small\textbf{Impact of diversity regulariser on OOD detection}. We show the model answer diversity, measured by PDS, and the OOD detection performance, measured by AUROC, against $\lambda$ values, the loss weight for the disagreement regularizer term.}
    \label{fig:lambda_ood_det}
    \vspace{0em}
\end{figure}

\textbf{Influence of diversification strength ($\lambda$).}
We further study the impact of ensemble diversification on the OOD detection with the PDS estimator. In Figure~\ref{fig:lambda_ood_det}, we observe that strengthening the diversification objective (higher $\lambda$) indeed leads to greater diversity (higher PDS), with a jump at around $\lambda\in[10^{-1},10^1]$. This range corresponds to the jump in the OOD detection performance (higher AUROC). 

\begin{wrapfigure}{r}{0.5\linewidth}
    \centering
    \small
    \includegraphics[width=1\linewidth]{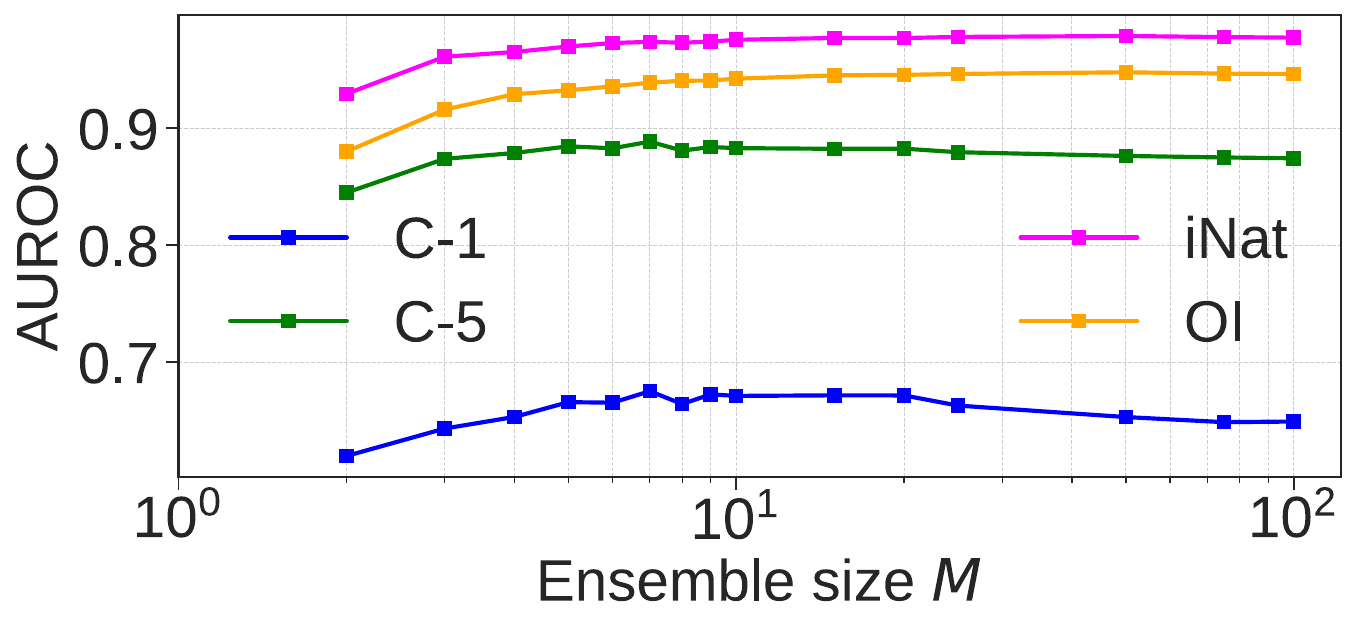}
    \vspace{-2em}
    \caption{\small\textbf{Impact of ensemble size on OOD detection}.}
    \label{fig:num_models}
\end{wrapfigure}

\textbf{Influence of ensemble size.}
How ensemble size influences performance of our method?
We can see that increasing ensemble size helps to improve AUROC for OOD detection on IN-C-1 (Figure~\ref{fig:num_models}).
Increasing ensemble size marginally helps, but using 5 models provides already a significant improvement over the smallest possible ensemble of size 2. It is also important to mention, that \ourmethodname{} framework is computationally more efficient \wrt ensemble size $M$ than A2D and DivDis: since we train ensembles for the fixed number of epochs, training complexity for \ourmethodname{} is $O(1)$ thanks to stochastic model pairs selection, while for A2D and DivDis it is $O(M^2)$.

\section{Related Work}

\label{sec:ensembles_general}
\textbf{Ensembling} is a well-known technique that aggregates the outputs of multiple models to make more accurate predictions~\citep{Breiman1996, Breiman2001, Hansen1990, Ho1995, Krogh1994}.
It is well known that diversity in the outputs of the ensemble members leads to better gains in performance~\citep{Krogh1994} because they make independent errors~\citep{DLbook, Hansen1990}.

In addition, it has been shown empirically and theoretically~\citep{yong2024spurious, hao2024benefits} that diverse ensembles can also improve OOD generalization. 

\label{sec:ensembles_regularizers}
\textbf{Diversity through regularizers.}
Various auxiliary training objectives have been proposed to encourage diversity across
models' weights~\citep{d2021repulsive, de2023maximum, demathelin2023deep, pmlr-v216-wang23c}, features~\citep{chen2024project, yashima2022feature, yong2024spurious}, input gradients~\citep{ross2020ensembles, Damien1, Damien2, trinh2024input}, or outputs~\citep{d2021repulsive, DivDis, liu1999simultaneous, A2D, scimeca2023shortcut}.
\citet{d2021repulsive} showed that a regularizer that repulses the ensemble members' weights or outputs leads to ensembles with a better approximation of Bayesian model averaging. This idea was extended by repulsing features~\citep{yashima2022feature} and input gradients~\citep{trinh2024input}.
Since ensemble are most useful when the errors of its members are uncorrelated~\citep{Krogh1994}, the closest of the above objective is to diversify their outputs.
This cannot be guaranteed with other objectives such as weight diversity for example, since two models could implement the exact same function with different weights due to the many symmetries in the parameter space of neural networks.
For this reason, this paper focuses on methods for output-space diversification~\citep{DivDis, A2D}. These were also highlighted as state-of-the-art in a recent survey on diversification~\citep{benoit2024unraveling}.

\textbf{Diversity without modifying the training objective.}
\label{sec:ensembles_other}
The most straightforward way to obtain diverse models is to independently train them
with different seeds (Deep Ensembles~\citep{Lakshminarayanan2017} and Bayesian extensions~\citep{MultiSwag2021}),
hyperparameters~\citep{wenzel2020hyperparameter},
augmentations~\citep{li2023whac}, or architectures~\citep{zaidi2021neural}.
A computationally cheaper approach is to use models saved at different points during the training~\citep{huang2017snapshot}
or models derived from the base model by applying dropout~\citep{gal2016dropout} or masking~\citep{durasov2021masksembles}.
The ``mixture of experts'' paradigm~\citep{DivE2} can also be viewed as an ensemble
where diversification happens by assigning different training samples to different ensemble members.
Our experiments use Deep Ensembles~\citep{Lakshminarayanan2017} and ensembles of models trained with different hyperparameters~\citep{wenzel2020hyperparameter}
as baselines since they are strong approaches to OOD detection~\citep{ovadia2019can} and OOD generalization especially when combined with ``model soups''~\citep{soup22}.

\section{Conclusions}
\label{sec:conclusion}

Ensemble diversification has many implications for treating one of the ultimate goals of machine learning, handling out-of-distribution (OOD) samples.
Training a large number of diverse hypotheses on a dataset is a way to generate
candidates that may have the desired OOD behaviour (i.e. better OOD generalization).
And the diversity of hypotheses can help distinguish ID from OOD samples by measuring  disagreements across ensemble members.
Despite these benefits, diverse-ensemble training has previously remained a lab-bound concept for two reasons. Previous approaches were computationally expensive (scaling quadratically with ensemble size) and required a separate OOD dataset to nurture the diverse hypotheses.

We have addressed these challenges through the novel Scalable Ensemble Diversification (SED) method. SED identifies OOD-like samples from the training data, bypassing the need to prepare a separate OOD data. SED also employs a stochastic pair selection to reduce the quadratic complexity of previous approaches to a constant one. We have demonstrated good performance of SED on OOD generalization and detection tasks, both at the ImageNet scale, a largely underexplored regime in the ensemble diversification community. In particular, for OOD detection, our novel diversity measure of Predictive Diversity Score (PDS) amplifies the benefits of diverse ensembles for OOD detection.

\textbf{Limitations.}~~
This work has focused on solving the applicability of disagreement-based diversification on realistic datasets. The contributions are thus mostly in the implementation, and the results focus on empirical benefits.
Work is needed to examine theoretical justifications for the method and characterize the exact conditions under which it should provide benefits.

Similarly, the proposed PDS is a conceptually sound measure of epistemic uncertainty,
but work is also needed to characterize the exact conditions where it is practically superior to other baselines.

\textbf{Acknowledgments.}~~~
This work was supported by the Tübingen AI Center. The authors thank the International Max Planck Research School for Intelligent Systems (IMPRS-IS) for supporting Alexander Rubinstein. Luca Scimeca acknowledges funding by Recursion Pharmaceuticals. This research utilized compute resources at the Tübingen Machine Learning Cloud, DFG FKZ INST 37/1057-1 FUGG.

{
    \small
    \bibliographystyle{abbrvnat}
    \bibliography{biblio}
}

\appendix
\section{Appendices}

\subsection{Varying the Number of Trainable Layers}

To perform an ablation study on the number of layers diversified for each ensemble member we trained only one last layer of DeiT-3b and compared it to the ensemble from the main experiments with the last two layers trained. Both ensembles have size 5 and were trained on the ImageNet training split. The results can be seen in Table~\ref{tab:layers}. Generalization performance did not change much, with the biggest change for ImageNet-C with the corruption strength 5 where ensemble accuracy dropped from $40.8\%$ for one layer to $40.6\%$ for two layers. However, OOD detection performance is better across the board for the case when two layers are diversified, for example, the detection AUROC scores for one layer diversified vs two layers diversified are $0.928$ vs $0.941$ for OpenImages and $0.964$ vs $0.977$ for iNaturalist. We believe that it can be explained by the fact that when one linear layer is trained with cross-entropy loss the optimization problem becomes convex making it harder for disagreement regularizer to promote diversity for different solutions, i.e. ensemble members tend to have similar weight matrices and disagree on OOD samples less.

\begin{table}[h!]
    \setlength{\tabcolsep}{.6em}
    \centering 
    \vspace{1em}
    \begin{tabular}{c|ccccc|cccc}
      \toprule
      & \multicolumn{5}{c}{Ensemble Acc.} & \multicolumn{4}{c}{AUROC} \\
      \# Layers & Val & IN-A & IN-R & C-1 & C-5 & C-1 & C-5 & iNat & OI \\
      \midrule
      1 & 85.2 & 42.3 & \textbf{48.2} & \textbf{77.3} & \textbf{40.8} & 0.677 & 0.889 & 0.964 & 0.928 \\
      2 & \textbf{85.3} & \textbf{42.4} & 48.1 & \textbf{77.3} & 40.6 & \textbf{0.681} & \textbf{0.894} & \textbf{0.977} & \textbf{0.941} \\
      \bottomrule
    \end{tabular}
    \vspace{.5em}
    \caption{Varying the number of trainable layers.\label{tab:layers}}
\end{table}

\subsection{Other Backbones}

To check the applicability of our method to other architectures we trained an ensemble of 5 models with the whole model but last layer frozen using ResNet18 as a feature extractor. We compared \ourmethodname{} with A2D disagreement regularizer and stochastic sum size $\mathcal{I}=2$ vs deep ensemble in Table~\ref{tab:rn18_backbone}. Both ensembles were trained on the ImageNet training split. Deep ensemble and \ourmethodname{-A2D } have similar generalization performance, with the biggest difference for ImageNet-C with the corruption strength 1 where ensemble accuracy dropped from $51.9\%$ for deep ensemble to $51.8\%$ for \ourmethodname{-A2D }. Nevertheless, \ourmethodname{-A2D } shows better OOD detection performance across the board, for example, the detection AUROC scores for one deep ensemble vs \ourmethodname{-A2D} are $0.802$ vs $0.812$ for OpenImages and $0.865$ vs $0.973$ for iNaturalist. Ensemble accuracy on ImageNet-A is less than $1\%$ for both ensembles: $0.5\%$ and $0.6\%$ because this dataset was created with a goal to minimize ResNet performance on it.

\begin{table}[h!]
    \centering
    \vspace{1em}
    \setlength{\tabcolsep}{.6em}
    \centering
    \small
    \begin{tabular}{l|ccccc|cccc}
     \toprule
    & \multicolumn{5}{c}{Ensemble Acc.} & \multicolumn{4}{c}{AUROC}\\    
     Method & Val & IN-A & IN-R & C-1 & C-5 & C-1& C-5& iNat& OI\\
     \midrule
     Deep Ensemble& \textbf{69.8}& 0.5& \textbf{20.8}& \textbf{51.9}& \textbf{14.6}& 0.670& 0.869& 0.865& 0.802\\
    \ourmethodname{-A2D}& 69.6& 0.6& \textbf{20.8}& 51.8& \textbf{14.6}& \textbf{0.686}& \textbf{0.879}& \textbf{0.873}& \textbf{0.812}\\
    \bottomrule
    \end{tabular}
    \vspace{.5em}
    \caption{{With a ResNet18 backbone.}}
    \label{tab:rn18_backbone}
\end{table}
    
\subsection{Other Uncertainty Scores}

To perform an ablation study on the OOD detection methods we compared PDS to other uncertainty scores computed for the outputs of \ourmethodname{} with A2D disagreement regularizer trained on the ImageNet training split. The ensemble size is 5, stochastic sum size $\mathcal{I}=2$. Results can be seen in the Table~\ref{tab:ood_scores}. In addition to popular baselines \citep{liu2020energy, xia2022usefulness}, we also used A2D disagreement regularizer as an uncertainty score (A2D-score in the table). PDS performed on par or better than other methods, for example, the biggest gap is achieved on OpenImages with detection AUROC score $0.941$ vs $0.917$ against A2D-score, and the smallest gap is achieved on iNaturalist with detection AUROC of $0.977$ for both PDS and Average Energy.

\begin{table}[h!]
\centering
      \setlength{\tabcolsep}{.9em}

\begin{tabular}{l c c c c}
                   \toprule
                   & C-1 & C-5 & iNat  & OI     \\
                   \midrule
BMA                & 0.641   & 0.845   & 0.960 & 0.915 \\
PDS                & \textbf{0.686}   & \textbf{0.896}   & \textbf{0.977} & \textbf{0.941} \\
A2D-score         & 0.685   & \textbf{0.896}   & 0.962 & 0.917 \\
Average Energy     & 0.633   & 0.858   & \textbf{0.977} & 0.908 \\
Average Entropy    & 0.580   & 0.825   & 0.960 & 0.916 \\
Average Max Prob   & 0.673   & 0.874   & 0.809 & 0.829 \\
Ens. Entropy       & 0.580   & 0.826   & 0.960 & 0.916 \\
Mutual information & 0.503   & 0.539   & 0.586 & 0.576 \\
\bottomrule
\end{tabular}
\vspace{.5em}
\caption{Different uncertainty scores used for OOD detection.}
\label{tab:ood_scores}
\end{table}

\subsection{Comparison to a Two-Stage Approach}

To perform an ablation study on the way samples for disagreement are selected in Table~\ref{tab:2-stage} we compared an ensemble trained with Equation~\ref{eq:sed} (called "joint" in the table) against a 2-stage approach. Instead of disagreeing on all samples with adaptive weight $\alpha_n$ as in Equation~\ref{eq:sed} we first computed the confidence of the pre-trained DeiT-3B model on all samples in ImageNet training split and then selected samples with a confidence lower than $0.2$ which resulted in $18002$ samples (to approximately match the sizes of ImageNet-A and ImageNet-R). Then we trained an ensemble by minimizing A2D disagreement regularizer on these samples while minimizing cross-entropy on all other samples. Both ensembles had size 5 and stochastic sum size $\mathcal{I}=2$.
While such an approach might sound simpler, \ourmethodname{} is more straightforward and efficient, since there is no need to train an initial model to determine samples for disagreement. Both ensembles have a similar generalization performance, with the biggest difference for ImageNet-R where ensemble accuracy dropped from $48.5\%$ for 2-stage approach to $48.1\%$ for the joint. In contrast, OOD detection performance is significantly better across the board for the joint approach, for example, the detection AUROC scores are $0.845$ vs $0.896$ for ImageNet-C with corruption strength 5 and $0.911$ vs $0.941$ for OpenImages. We think that such a drastic difference in OOD detection performance can be caused by the fact that the set of samples selected for disagreement may be suboptimal which makes the confidence threshold (set as $0.2$ for this experiment) an important hyperparameter and adds even more complexity to the 2-stage approach.

\begin{table}[h!]
    \centering
    \vspace{1em}
      \setlength{\tabcolsep}{.5em}
      \centering
         \begin{tabular}{l|ccccc|cccc}
         \toprule        
        & \multicolumn{5}{c}{Ensemble Acc.} & \multicolumn{4}{c}{AUROC}\\
         \midrule        
         Type& Val & IN-A & IN-R & C-1 & C-5 & C-1& C-5& iNat& OI\\
         \midrule
         2-stage& 85.2& \textbf{42.4}& \textbf{48.5}& \textbf{77.3}& \textbf{40.7}& 0.597& 0.845& 0.960& 0.911\\
     Joint& \textbf{85.3}& \textbf{42.4}& 48.1& \textbf{77.3}& 40.6& \textbf{0.686}& \textbf{0.896}& \textbf{0.977}& \textbf{0.941}\\
     \bottomrule
     \end{tabular}
     \vspace{.5em}
     \caption{{Comparison with a two-stage approach.}}
     \label{tab:2-stage}
\end{table}

\subsection{Small-Scale Experiments}

To check the performance of our method on the small-scale datasets, we conducted additional experiments on the Waterbirds dataset \citep{Sagawa*2020Distributionally} (Table~\ref{tab:waterbirds}), since both A2D and DivDis also provided results on it. We report the worst group (test) accuracy for ensembles of size 4. We trained A2D, DivDis, and an ensemble with \ourmethodname{} and A2D disagreement regularizer on Waterbirds training split. We did not use stochastic sum for \ourmethodname{-A2D} to factor out its influence. A2D and DivDis used the validation set for disagreement. While DivDis discovers a better single model having best accuracy of $87.2\%$ against $83.2\%$ for the proposed \ourmethodname{-A2D} method, the ensemble is clearly better with \ourmethodname{-A2D}: $80.6\%$ vs $78.3\%$ for DivDis. 

\begin{table}[h!]
\centering
\setlength{\tabcolsep}{.7em}
\begin{tabular}{l c c}
\toprule
       & Oracle selection & Ensemble  \\
\midrule
ERM    & 76.5   & 72.0      \\
DivDis & \textbf{87.2}   & 78.3      \\
A2D    & 78.3   & 78.3      \\
SED    & 83.5   & \textbf{80.6}     \\
\bottomrule
\end{tabular}
\vspace{.5em}
\caption{Worst group test accuracy on Waterbirds}
\label{tab:waterbirds}
\end{table}

\subsection{OOD Datasets for Disagreement}

To analyze the influence of OOD data used for disagreement we performed additional experiments with ensemble members disagreeing on ImageNet-R and ImageNet-A in Table~\ref{tab:ood_abl}. We compare A2D and Div \citep{DivDis} diversification regularizers.
 Usage of ImageNet-A or ImageNet-R resulted in almost identical (identical after rounding) OOD generalization performance for A2D disagreement regularizer, while for Div regularizer ensemble accuracy on ImageNet-R dropped from $45.2\%$ when using ImageNet-A for disagreement to $41.8\%$ when using ImageNet-R for disagreement. OOD detection performance also does not change much for any combination of regularizer and dataset used for disagreement with the biggest difference in detection AUROC scores $0.973$ for Div regularizer and and ImageNet-A disagreement dataset vs $0.969$ for Div regularizer and ImageNet-R disagreement dataset.  
 
\begin{table}[h!]
        \setlength{\tabcolsep}{.5em}
        \vspace{1em}
    \centering
    \begin{tabular}{ll|ccccc|cccc}
      \toprule
      && \multicolumn{5}{c}{Ensemble Acc.} & \multicolumn{4}{c}{AUROC} \\
      Method&OOD& Val & IN-A & IN-R & C-1 & C-5 & C-1& C-5& iNat& OI \\
      \midrule
      A2D&IN-A& \textbf{85.1}& \textbf{37.8}& \textbf{45.2}& \textbf{77.2}& \textbf{40.3}& 0.599& \textbf{0.850}& 0.971& 0.936 \\
      A2D&IN-R& \textbf{85.1}& \textbf{37.8}& \textbf{45.2}& \textbf{77.2}& \textbf{40.3}& 0.599& \textbf{0.850}& 0.971& \textbf{0.939} \\
      Div&IN-A& \textbf{85.1}& \textbf{37.8}& \textbf{45.2}& \textbf{77.2}& \textbf{40.3}& 0.599& \textbf{0.850}& \textbf{0.973}&0.937 \\
      Div& IN-R& \textbf{85.1}& 35.7& 41.8& \textbf{77.2}& 40.2& \textbf{0.600}& \textbf{0.850}& 0.969&0.938 \\
      \bottomrule
    \end{tabular}
    \vspace{.5em}
    \caption{{OOD Datasets for disagreement.}}
    \label{tab:ood_abl}
\end{table}

\subsection{Variations of the Stochastic Sum Size}

We performed an additional evaluation (Table~\ref{tab:stoch_sum_abl}) that shows the benefit of controlling the stochastic sum size ($\mathcal{I}$) on the speed of training an ensemble. For example, to train an ensemble of size 5, the time required for 1 epoch grows from 53s for $\mathcal{I}=2$ to 585s for $\mathcal{I}=5$ (without stochastic sum). We could not train an ensemble of 50 models without stochastic sum with our resources, but it already requires 7244s for $\mathcal{I}=10$ vs 2189s for $\mathcal{I}=2$. Standard deviations of training epoch times are computed across 10 different epochs.

\begin{table}[h!]
    \setlength{\tabcolsep}{.55em}
    \centering
    \vspace{1em}
    \begin{tabular}{ll|l|ccccc|cccc}
      \toprule
      &&  &\multicolumn{5}{c}{Ensemble Acc.} & \multicolumn{4}{c}{AUROC} \\
      M&I&  Epoch, s&Val & IN-A & IN-R & C-1 & C-5 & C-1& C-5& iNat& OI \\
      \midrule
      5&2&  \textbf{53 $\pm$ 5}&\textbf{85.3}& \textbf{42.4}& \textbf{48.1}& \textbf{77.3}& \textbf{40.6}& 0.686& 0.896& \textbf{0.977}& \textbf{0.941} \\
      5&3&  388 $\pm$ 28&85.2& 41.4& 47.4& 77.2& 40.5& 0.682& 0.892& 0.975& 0.939 \\
      5&4&  423 $\pm$ 3&85.2& 40.3& 46.8& 77.1& 40.4& 0.703& 0.898& 0.973&0.940 \\
      5&5&  585 $\pm$ 111&85.1& 37.6& 44.9& 77.0& 40.2& \textbf{0.711}& \textbf{0.903}& 0.970&0.937 \\
      \midrule
      50&2&  \textbf{2189 $\pm$ 86}&\textbf{83.7}& \textbf{50.1}& \textbf{54.0}& \textbf{75.9}& \textbf{39.4}& \textbf{0.600}& 0.824& 0.934&0.878 \\
      50&5&  4213 $\pm$ 5&83.6& 49.2& 53.4& 75.8& 39.2& 0.598& 0.827& 0.942&0.892 \\
      50&10&  7244 $\pm$ 27&83.4& 48.5& 53.0& 75.6& 39.1& 0.597& \textbf{0.828}& \textbf{0.945}&\textbf{0.896} \\
      \bottomrule
    \end{tabular}
    \vspace{.5em}
    \caption{{Variations of the stochastic sum size.}}
    \label{tab:stoch_sum_abl}
\end{table}

\end{document}